\documentclass{isprs} 
\usepackage{subfigure}
\usepackage{setspace}
\usepackage{geometry} 
\usepackage{epstopdf}
\usepackage[labelsep=period]{caption}  
\usepackage[british]{babel} 
\usepackage[hang]{footmisc}
\usepackage{multicol}
\usepackage{graphicx}        
\usepackage{subcaption}      
\usepackage{caption}         
\usepackage{float} 
\usepackage{wrapfig}
\usepackage{natbib}
\usepackage{hyperref}
\usepackage{url}
\usepackage{enumitem}

\usepackage{amsmath,amsfonts,bm}

\def\eqref#1{equation~\ref{#1}}

\def\1{\bm{1}}

\def\vc{{\bm{c}}}

\def\vp{{\bm{p}}}

\def\vv{{\bm{v}}}

\def\mA{{\bm{A}}}

\def\mC{{\bm{C}}}

\DeclareMathAlphabet{\mathsfit}{\encodingdefault}{\sfdefault}{m}{sl}
\SetMathAlphabet{\mathsfit}{bold}{\encodingdefault}{\sfdefault}{bx}{n}

\begin{document}

\title{An Interpretable Implicit-Based Approach for Modeling Local Spatial Effects: A Case Study of Global Gross Primary Productivity Estimation
}
\date{}

\author{
 Siqi Du\textsuperscript{1} \thanks{Corresponding author}, Hongsheng Huang\textsuperscript{2}, Kaixin Shen\textsuperscript{3}, Ziqi Liu\textsuperscript{4}, Shengjun Tang\textsuperscript{5}}

\address{
	\textsuperscript{1 }College of Urban and Environmental Sciences, Peking University, Beijing, P.R. China\\
        \textsuperscript{2 }Department of Land Surveying and Geo-Informatics, The Hong Kong Polytechnic University, Hong Kong, P.R. China\\
	\textsuperscript{3 }School of Safety Science, Tsinghua University, Beijing, P.R. China\\
	\textsuperscript{4 }College of Surveying and Geo-Informatics, Tongji University, Shanghai, P.R. China\\
        \textsuperscript{5 }Research Institute for Smart Cities, School of Architecture and Urban Planning, Shenzhen University, Shenzhen, P.R. China\\
}

\abstract{

In Earth sciences, unobserved factors exhibit non-stationary spatial distributions, causing the relationships between features and targets to display spatial heterogeneity. In geographic machine learning tasks, conventional statistical learning methods often struggle to capture spatial heterogeneity, leading to unsatisfactory prediction accuracy and unreliable interpretability. While approaches like Geographically Weighted Regression (GWR) capture local variations, they fall short of uncovering global patterns and tracking the continuous evolution of spatial heterogeneity. Motivated by this limitation, we propose a novel perspective—that is, simultaneously modeling common features across different locations alongside spatial differences using deep neural networks. The proposed method is a dual-branch neural network with an encoder-decoder structure. In the encoding stage, the method aggregates node information in a spatiotemporal conditional graph using GCN and LSTM, encoding location-specific spatiotemporal heterogeneity as an implicit conditional vector. Additionally, a self-attention-based encoder is used to extract location-invariant common features from the data. In the decoding stage, the approach employs a conditional generation strategy that predicts response variables and interpretative weights based on data features under spatiotemporal conditions. The approach is validated by predicting vegetation gross primary productivity (GPP) using global climate and land cover data from 2001 to 2020. Trained on 50 million samples and tested on 2.8 million, the proposed model achieves an RMSE of 0.836, outperforming LightGBM (1.063) and TabNet (0.944). Visualization analyses indicate that our method can reveal the distribution differences of the dominant factors of GPP across various times and locations.
}


\maketitle


\section{Introduction}\label{Manuscxript}
 In Earth science, machine learning method like XGBoost \citep{chen2016xgboost} is widely used to model environmental and geographical relationships, such as predicting climate change impacts on vegetation \citep{lu2024global} and understanding tropical cyclones' effects on precipitation \citep{qin2024global}. However, a fundamental challenge in geographical modeling is the presence of spatiotemporal heterogeneity, where relationships between independent and target variables vary across both space and time. Most machine learning methods, which assume unordered samples and stationary relationships, lack the capability to model such spatiotemporal heterogeneity. This limitation calls for novel approaches specifically designed for geographical machine learning problems that can capture these varying relationships while maintaining model accuracy and interpretability.

One solution is to fit, at different spatial locations, a set of weights that vary with the coordinates by locally sampling the data, with Geographically Weighted Regression (GWR) \citep{fotheringham2009geographically} serving as a representative method. However, GWR lacks temporal modeling, limiting its ability to capture the evolution of spatial heterogeneity. To address GWR's temporal limitation, Geographically and Temporally Weighted Regression (GTWR) \citep{fotheringham2015geographical}  was developed, using spatiotemporal metrics to model variability. Despite these advancements, current methods still fit spatial weights locally, based on neighborhood samples, leading to several challenges: 1) Local models capture spatial variability but often overlook the shared global patterns. 2) Many spatial models depend on heuristic parameter tuning, lacking automatic optimization of spatial weights. 3) Training separate models for each location is computationally inefficient for large-scale Earth science data.

\begin{figure*}[ht] 
    \centering  
    \includegraphics[width=0.9\textwidth]{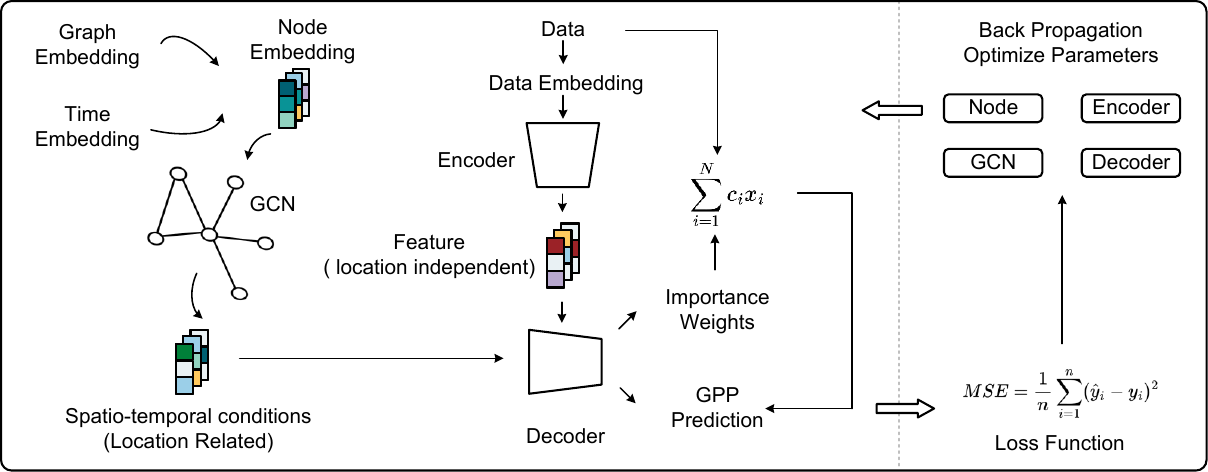}  
    \caption{The overall process of the method.}  
    \label{fig:overall} 
\end{figure*}  

In response to these challenges, we review the inherent characteristics of geographic machine learning tasks. Compared to typical machine learning problems, geographic machine learning tasks generally require the simultaneous capture of global common patterns and local spatial differences. For instance, when investigating the relationship between vegetation gross primary productivity (GPP) and various influencing factors. Although vegetation growth is governed by a general law, the dominant factors vary slightly in different locations. Based on this observation, our study proposes a novel perspective: \textbf{the geographic machine learning problem can be decomposed into two sub-tasks—learning the location-invariant objective law and capturing the conditional differences among various spatial locations.}

Based on the considerations above, we propose a novel geographic machine learning approach. This method is inspired by deep learning research that emphasizes direct pattern discovery from data and end-to-end differentiable processes for the simultaneous optimization of multiple tasks \citep{carion2020end}. The proposed approach leverages a multi-branch deep neural network to concurrently model common features shared across different locations and local spatial differences. It employs an encoder-decoder architecture. In the encoding stage, node information is aggregated in a spatiotemporal conditional graph using graph convolutional networks (GCN) and long short-term memory (LSTM) networks to encode location-specific spatiotemporal heterogeneity as an implicit conditional vector. Additionally, a self-attention-based encoder is utilized to extract location-invariant common features from the data. In the decoding stage, a conditional generation strategy is adopted to predict the response variables and interpretative weights based on data features under spatiotemporal conditions.

To validate our approach, we created the Climate2GPP dataset, using the ERA5 climate dataset \citep{munoz2021era5}, the MCD12C1.061 MODIS Land Cover dataset \citep{friedl_sulla-menashe_2022}, and the PML\_V2 0.1.7 GPP dataset \citep{zhang2019coupled}. Spanning from 2001 to 2020 with an 8-day temporal resolution, this dataset includes approximately 50 million samples for training and 2.8 million samples for testing. Our method achieved an RMSE of 0.836 on the test set, significantly outperforming GWR (RMSE 1.937), classical tabular machine learning methods like LightGBM Large (RMSE 1.063) and deep learning methods like TabNet (RMSE 0.944). 

Our main contributions are summarized as follows:
\begin{itemize} 
\item[$\bullet$] A novel multi-branch encoder-decoder architecture that \\ jointly models global features and spatial differences.
\item[$\bullet$] An integrated spatiotemporal heterogeneity model using conditional graphs, GCN, and LSTM.
\item[$\bullet$] A dual-branch decoder with cross-attention for effective fusion and prediction under spatiotemporal constraints.
\end{itemize}

\section{Related Works}
Tabular machine learning methods have been widely applied in Earth science for tasks such as predicting environmental changes and understanding geographical phenomena. These methods, including popular algorithms like LightGBM \citep{ke2017lightgbm} and XGBoost \citep{chen2016xgboost}, are designed under the assumption that samples are independent and identically distributed, which limits their applicability in scenarios with spatial dependencies. While geographically weighted models, such as GWR \citep{fotheringham2009geographically} and GTWR \citep{fotheringham2015geographical}, have been introduced to address spatial heterogeneity by adjusting coefficients locally, they face significant limitations. GWR models fail to capture temporal evolution, and while GTWR extends this capability, both models struggle with nonlinearity and exhibit high computational complexity when applied to large datasets. GWR-RF \citep{wang2024geographically} combines GWR with Random Forest for nonlinearity but still suffers from local overfitting and dense weight matrices. GNNWR \citep{du2020geographically} balances global patterns and spatial variability through neural network-corrected coefficients but retains the complexity of dense spatial weights. GTNNWR \citep{wu2021geographically} further incorporates spatiotemporal heterogeneity but remains limited by the need to fit local variables and dense spatiotemporal weights, making these models prone to overfitting and computationally inefficient in large-scale applications.

\section{Method}
The proposed method is built upon a multi-branch architecture with an encoder-decoder structure that jointly addresses two key objectives: extracting location invariant common features and learning location specific spatiotemporal differential conditions. In the encoding stage, one branch utilizes a dual self-attention mechanism within a tabular data encoder to capture shared, common features from data. Concurrently, another branch focuses on learning spatiotemporal differential conditions, representing location specific variations via implicit latent vectors. These latent representations are derived through a locally shared spatiotemporal condition graph and further refined using Graph Convolutional Networks (GCN) and Long Short-Term Memory (LSTM) networks.

In the decoding stage, a multi-branch decoder integrates the location-invariant features with the spatiotemporal conditions through cross-attention interactions. This decoder is designed under two distinct objectives: predicting the target variable and estimating feature importance weights. The architecture, divided into a target variable prediction branch and a feature importance estimation branch, ensures that predictions are made under the appropriate spatiotemporal constraints while maintaining model interpretability. Finally, all parameters are jointly optimized under both the main loss and the auxiliary loss. \\ \\

\begin{figure*}[ht] 
    \centering  
    \includegraphics[width=0.94\textwidth]{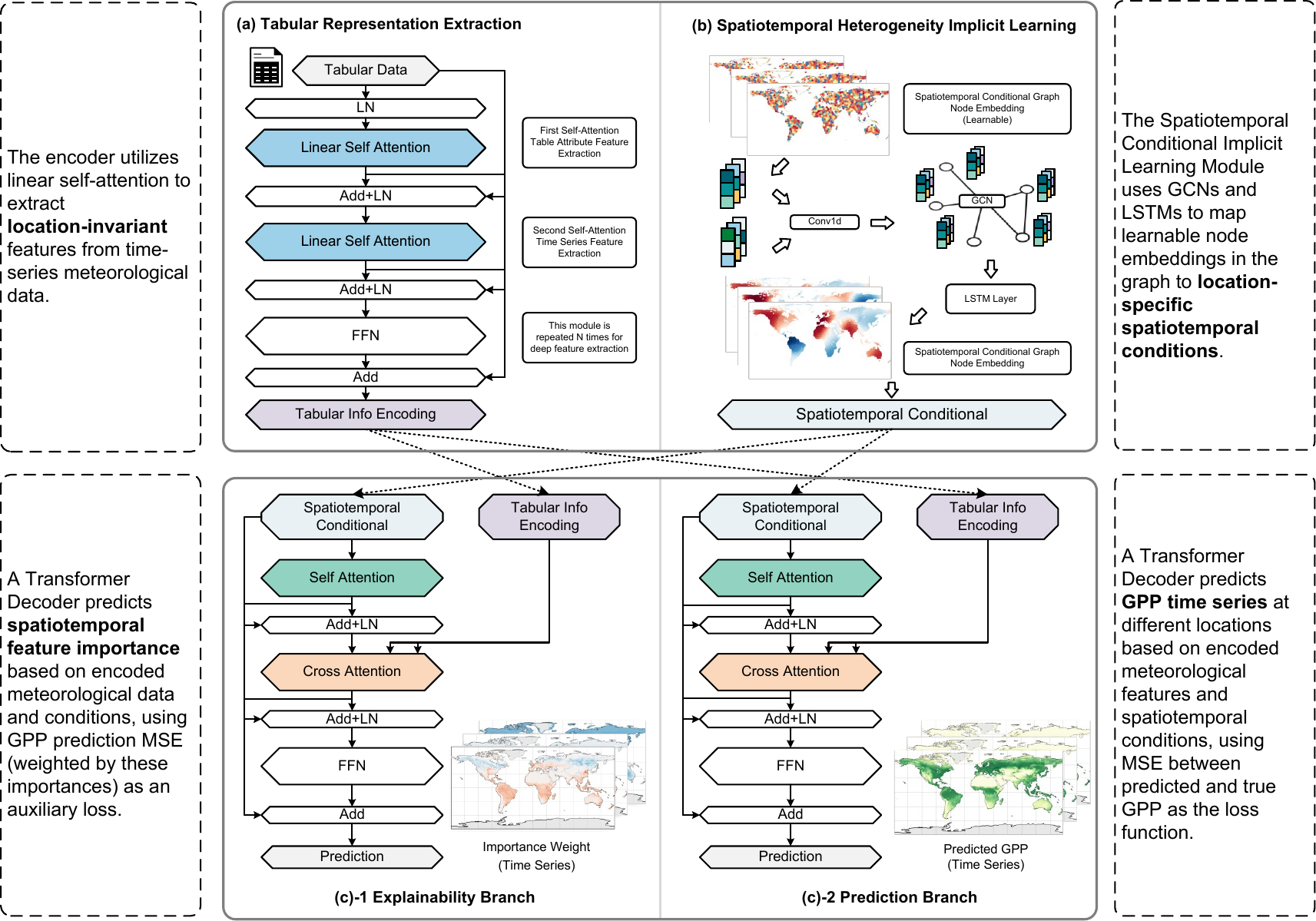}  
    \caption{The implementation details of the dual encoder and the dual decoder.}  
    \label{fig:overall} 
\end{figure*}

\subsection{Extraction of Location-Invariant Common Features}
In our task, the extraction of location-invariant common features means learning robust representations from time series meteorological data. To achieve this, we designed a module that decomposes the feature extraction task into two complementary objectives: (1) capturing inter-feature relationships among the data attributes, and (2) extracting temporal patterns inherent in the time series of each attribute.

To address these objectives, we extend a transformer encoder originally developed for tabular data \citep{hu2024pytorch}. This enhanced encoder, as illustrated in Figure 2(a), incorporates a dual attention (DA) mechanism that simultaneously operates in both feature and temporal dimensions. To further accommodate the large-scale geoscience data and reduce the computational complexity, we implement a novel linear self-attention mechanism \citep{pmlr-v119-katharopoulos20a}. After performing linear embedding and applying temporal position encoding \citep{su2024roformer} to the input meteorological data, the stacked DA modules progressively refine the representations, ultimately yielding data features that are invariant to location-specific factors.

\textbf{Dual Attention Mechanism:} Given an input tensor \( X \in \mathbb{R}^{L \times D} \), where \( L \) represents the sequence length (time dimension) and \( D \) is the feature dimension, the DA module sequentially computes self-attention \citep{pmlr-v119-katharopoulos20a} across the temporal and feature dimensions. First, temporal self-attention is applied across the time steps for each feature. Queries, keys, and values are computed as:
\begin{equation}
Q^{\text{temp}} = X W_Q^{\text{temp}}, \quad K^{\text{temp}} = X W_K^{\text{temp}}, \quad V^{\text{temp}} = X W_V^{\text{temp}}
\end{equation}
The temporal attention output is then computed by applying the activation function \( \phi(x) = \text{ELU}(x) + 1 \) directly within the attention formula:
\begin{equation}
\text{Attn}^{\text{temp}} = \frac{\phi(Q^{\text{temp}}) \left( \phi(K^{\text{temp}})^\top V^{\text{temp}} \right)}{\phi(Q^{\text{temp}}) \left( \phi(K^{\text{temp}})^\top \mathbf{1}_L \right) + \epsilon}
\end{equation}
Feature self-attention is then computed along the feature dimension using the same process. Residual connections are employed at each step to ensure gradient flow and model stability:
\begin{equation}
X^{\text{feat}} = X^{\text{temp}} + \text{Attn}^{\text{feat}}
\end{equation}
\textbf{Feedforward Network:} Finally, a position-wise feedforward network is applied to each element:
\begin{equation}
Y = X^{\text{feat}} + \text{FFN}(X^{\text{feat}})
\end{equation}
By employing both temporal and feature self-attention, followed by a feedforward network, this model captures rich representations from tabular Earth science data. \\

\begin{figure*}[htbp] 
    \centering  
    \includegraphics[width=0.95\textwidth]{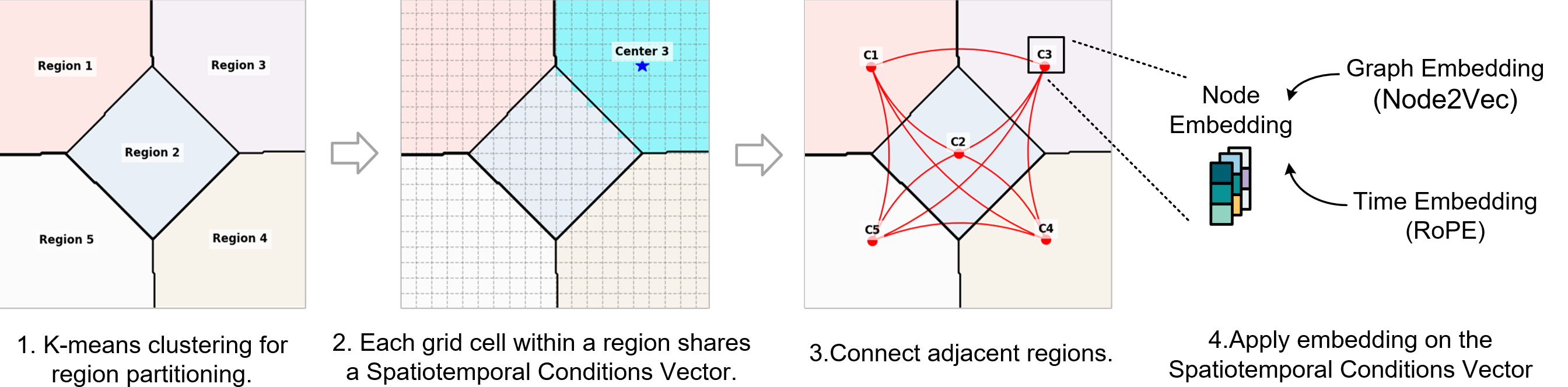}  
    \caption{Spatiotemporal Conditional Graph Construction with Local Weight Sharing.}
    \label{fig:STCG} 
\end{figure*}

\subsection{Learning Spatiotemporal Differential Conditions}
The Learning Spatiotemporal Differential Conditions subsection is dedicated to directly learning implicit condition vectors that capture the distinct spatiotemporal variations present across locations. To achieve this, each location is assigned a hidden vector representing its unique spatiotemporal condition. These vectors, together with their spatial relationships, form the Spatiotemporal Conditional Graph. By leveraging a combination of Graph Convolutional Networks (GCN) and Long Short-Term Memory (LSTM) networks, our approach effectively aggregates the spatial dependencies and temporal dynamics among these vectors, resulting in a unified spatiotemporal condition vector that is subsequently input into the decoder for further prediction tasks. \\

\textbf{Spatiotemporal Conditional Graph Construction (STCG): }The STCG is defined as \( G = (V, E) \), where \( V \) is the set of nodes and \( E \) is the set of edges. Each node \( v_{i,t} \in V \) represents a spatiotemporal point \( (\lambda_i, \phi_i, t) \), where \( \lambda_i \) and \( \phi_i \) are the longitude and latitude of node \( i \), and \( t \) is the time. Each node has an embedding \( \vv_{i,t} \) that captures the spatiotemporal condition at that point. The prediction for a spatiotemporal location is influenced by the embedding \( \vv_{i,t} \) of the corresponding node in the STCG. The construction of the STCG involves the following steps. The construction of STCG is shown in Figure \ref{STCG}

\textbf{Graph Node Generation:} To determine the geographical coordinates \( (\lambda_i, \phi_i) \) of each node \( v_{i,t} \in V \), we apply K-means clustering to the global land grid. This step reduces computational complexity by grouping spatial regions into clusters, where each cluster shares a common spatiotemporal condition. Specifically, the cluster centers \( \mC = \{\vc_1, \vc_2, \dots, \vc_k\} \) are determined by minimizing the sum of squared distances between all spatial points and their nearest cluster centers:
\begin{equation}
\mC = \arg \min_{\mC} \sum_{p=1}^{n} \min_{j} \| (\lambda_p, \phi_p) - \vc_j \|^2
\end{equation}
Here, \( (\lambda_p, \phi_p) \) represents any spatial point \( p \) in the global grid, and \( n \) is the total number of such points. Each node \( v_{i,t} \in V \) is then assigned the spatial coordinates of the corresponding cluster center: \((\lambda_i, \phi_i) = \vc_i\).

\textbf{Cyclic Graph Construction:} To ensure connectivity between the eastern and western hemispheres, we map the geographical coordinates of the cluster centers to spherical coordinates and construct the adjacency matrix \( \mA \) using the K-nearest neighbors (KNN) method on the sphere. Specifically, for each cluster center \( v_{i,t} \), we first project its geographical coordinates \( (\lambda_i, \phi_i) \) (longitude and latitude) onto a 3D unit sphere using the following transformation:
\begin{equation}
x_i = \cos(\phi_i) \cos(\lambda_i), \quad y_i = \cos(\phi_i) \sin(\lambda_i), \quad z_i = \sin(\phi_i)
\end{equation}
where \( (x_i, y_i, z_i) \) represents the 3D spherical coordinates of node \( v_{i,t} \). Using the spherical coordinates \( \vp_i = (x_i, y_i, z_i) \), we compute the adjacency matrix \( \mA \) of the graph by defining the \( k \)-nearest neighbors for each node \( v_{i,t} \). The adjacency matrix \( \mA \) is constructed as follows:
\begin{equation}
\mA_{i,j} = 
\begin{cases}
1, & j \in \operatorname{arg\,top-k} \| \vp_i - \vp_j \| \\
0, & \text{otherwise}
\end{cases}
\end{equation}
Here, \( \vp_i\) and \( \vp_j\) are the 3D spherical coordinates of nodes \( v_{i,t} \) and \( v_{j,t} \), respectively. This approach ensures that the graph is cyclic, connecting locations on opposite sides of the globe, which is particularly important for capturing the circular nature of the Earth.

\textbf{Node Embedding Calculation:} We use Node2vec \citep{grover2016node2vec} to compute the initial embeddings for the nodes. For each time dimension \( t \), we add a time embedding using the Rotational Position Embedding (RoPE) \citep{su2024roformer} method:
\(\vv_{i,t} = \text{Node2vec}(i) + \text{RoPE}(t)\)
where \( \vv_{i,t} \) is the embedding of node \( v_{i,t} \) at time \( t \).

\textbf{Edge Weight Calculation:} The weight of each edge \( w_{i,j} \) in the graph is computed using a log-Gaussian kernel, which incorporates two sequential normalization steps to effectively capture the similarity between cluster centers in a 3D space. The weight of each edge \( w_{i,j} \) can be calculated by:
\begin{equation}
w_{i,j} =
\begin{cases}
\exp\Biggl(-\dfrac{1}{2\sigma^2}
\Bigl(
1 - \exp\Bigl(-\dfrac{\|\vp_i-\vp_j\|}{\mu}\Bigr)
\Bigr)^2
\Biggr) \\[1ex]
0
\end{cases}
\end{equation}
The above equation defines $w_{i,j}$ such that when 
\begin{equation}
j \in \operatorname*{arg\,top-k}\|\mathbf{p}_i - \mathbf{p}_j\|,
\end{equation}
the weight is given by the expression in the first case; otherwise, it is 0.

\textbf{Spatiotemporal Conditional Encoding: } Although our earlier constructed spatiotemporal conditional graph enables each node to directly learn its spatiotemporal condition vector via back-propagation, this method alone is not sufficient to capture the intricate interactions between regions. To address this limitation, we incorporate a spatiotemporal aggregation operation that leverages 1D temporal convolution, graph convolution, and LSTM. For each node \( v_{i,t} \) in the STCG, we update its embedding \( \vv_{i,t} \) through these aggregation processes.

\textbf{Temporal Aggregation:} We first apply a 1D convolution operation along the time dimension to capture local temporal dependencies. This can be formulated as:
\begin{equation}
V^{temp} = V * W^{time}
\end{equation}
where \( V \) is the matrix of node embeddings \( \vv_{i,t} \), \( W^{time} \) is the learnable temporal convolution kernel, and \( * \) denotes the convolution operation.

\textbf{Spatial Aggregation:} Following the temporal aggregation, we employ a graph convolution operation to aggregate spatial information:
\begin{equation}
V^{spatial} = \sigma(D^{-\frac{1}{2}} \mA W D^{-\frac{1}{2}} V^{temp} H)
\end{equation}
where \( \mA \) is the adjacency matrix, \( W \) is the edge weight matrix \( w_{i,j} \), \( D \) is the degree matrix \( d_i \), \( H \) is the learnable weight matrix, and \( \sigma \) is a non-linear activation function. The final embedding for each node \( v_{i,t} \), incorporating both temporal and spatial information, can be expressed as:
\begin{equation}
\vv_{i,t}^{spatial} = \sigma\left(\sum_{j \in \mathcal{N}(i)} \frac{1}{\sqrt{d_i d_j}} w_{i,j} H \left(\sum_{\tau=-k}^{k} w_{\tau}^{time} \cdot \vv_{j,t+\tau}\right)\right)
\end{equation}
where \( \mathcal{N}(i) \) is the set of neighboring nodes of node \( v_{i,t} \), \( d_i \) and \( d_j \) are the degrees of nodes \( i \) and \( j \), \( w_{i,j} \) is the edge weight between nodes \( v_{i,t} \) and \( v_{j,t} \), and \( w_{\tau}^{time} \) are the elements of the temporal convolution kernel \( W^{time} \). By applying these operations sequentially, we obtain a rich representation \( \vv_{i,t}^{spatial} \) for each spatiotemporal point \( v_{i,t} \), which encapsulates both local and global spatiotemporal dependencies.

\textbf{LSTM Aggregation:} After performing spatial aggregation, these spatial features are further processed by an LSTM network to capture long-term temporal dependencies.

\begin{equation}
V^{final} = \text{LSTM}(V^{spatial})
\end{equation}

\subsection{Decoder Design for Constrained Prediction}
In the prediction stage, a conditional prediction approach is adopted to estimate the target variables corresponding to meteorological data under spatiotemporal conditions. The prediction can be formulated under the maximum a posteriori (MAP) framework as follows:
\begin{equation}
\hat{y} = \arg\max_{y} \; p\Big( y \,\big|\, c(x,y,t),\, X \Big)
\end{equation}
where \(c(x,y,t)\) denotes a function of spatial coordinates (\(x, y\)) and time (\(t\)) representing the spatiotemporal condition and \(X\) stands for the input data. To achieve a higher level of interpretability similar to geographically weighted regression (GWR), our design employs two independent branches: one for predicting the target variable and another for estimating the interpretable weights associated with the intarget variables.

\textbf{Decoder Structure:} \quad For the decoder, we utilize the standard Transformer decoder \citep{vaswani2017attention}, which naturally serves as a conditional predictor. Its core component is the cross-attention mechanism. In our framework, the query \(Q\) and key \(K\) are embedded from the data features obtained in the \textbf{Extraction of Location-Invariant Common Features}, while the value \(V\) is given by \(V^{final}\), which aggregates the spatiotemporal conditional graph information. The cross-attention mechanism is formulated as follows:
\begin{equation}
\text{Attention}(Q, K, V) = \text{softmax}\left(\frac{QK^{T}}{\sqrt{d_k}}\right)V,
\end{equation}
where \(d_k\) is the dimension of the key vectors.

\textbf{target variable Prediction Branch:} \quad In this branch, the Transformer decoder takes as input the spatiotemporal conditions and the encoded data features to directly predict the target variable. The optimization of this branch is performed using the Mean Squared Error (MSE) loss:
\begin{equation}
\mathcal{L}_{\text{dep}} = \frac{1}{N} \sum_{i=1}^{N} \Big(y_i - \hat{y}_i\Big)^2,
\end{equation}
where \(y_i\) and \(\hat{y}_i\) denote the ground-truth and predicted values, respectively.

\textbf{Feature Importance Estimation Branch:} \quad For the estimation of interpretable weights, we employ a separate Transformer decoder branch to predict an importance weight for each intarget variable. The predicted interpretable weights, denoted as \(w_j\), are then used to perform a weighted linear combination of the input intarget variables:
\begin{equation}
\hat{y}_{\text{interp}} = \sum_{j=1}^{M} w_j \, x_j,
\end{equation}
where \(x_j\) represents the \(j\)th intarget variable and \(M\) is the total number of intarget variables. An auxiliary MSE loss is computed between \(\hat{y}_{\text{interp}}\) and the true target variable \(y\) to further refine the interpretability of this branch:
\begin{equation}
\mathcal{L}_{\text{interp}} = \frac{1}{N} \sum_{i=1}^{N} \Big(y_i - \hat{y}_{\text{interp},i}\Big)^2.
\end{equation}

The overall training objective is to minimize a combination of both loss functions $\mathcal{L}_{\text{interp}}+\mathcal{L}_{\text{dep}}$, thereby ensuring accurate prediction while obtaining interpretable feature importance scores.

\section{Experiments}
\label{others}
\subsection{Experimental Setup}
\textbf{Dataset: }To validate our approach, we created the Climate2GPP dataset. We used data from Google Earth Engine spanning January 1, 2001, to December 17, 2020. The data sources include:
\begin{itemize}
    \item \textbf{ERA5-Land Daily Aggregated} (ECMWF): Global historical meteorological data aggregated every 8 days \citep{munoz2021era5}.
    \item \textbf{MCD12C1.061 MODIS Land Cover Type} (NASA): Yearly global land cover changes at 0.05 degree resolution \citep{friedl_sulla-menashe_2022}.
    \item \textbf{PML\_V2 0.1.7}: Global gross primary productivity (GPP) data aggregated every 8 days \citep{zhang2019coupled}.
\end{itemize}
From ERA5, we selected 26 climate parameters, with solar radiation, evaporation, and precipitation summed over 8-day periods, while the rest were averaged. GPP was similarly summed over 8-day intervals. Data from 2001 to 2019 was used for training (52M samples), and data from 2020 served as the test set (2.8M samples).

\textbf{Training Setting:}  
Our method is implemented in PyTorch 2.1.2 with CUDA 11.8. All features, except GPP, are normalized. The AdamW optimizer is used with a batch size of 256, an initial learning rate of 0.001, decayed to 0.0001 after 10 epochs, for a total of 20 epochs.

Comparison machine learning method is using AutoGluon 1.1.1 \citep{erickson2020autogluon} with default hyperparameters. These models, along with deep learning comparisons (TabNet, ResNet, ExcelFormer, FFTransformer implemented in PyTorchFrame 0.2.3 \citep{hu2024pytorch}), are trained on RTX 4090 GPU, 64-core Intel Xeon Platinum 8352V and 120GB of RAM. All deep learning models use the same optimizer settings as our method.

\subsection{Result Comparison}
\begin{table}[h]
    \centering
    \caption{Comparison of Different Methods}
    \label{tab:comparison_methods}
    \setlength{\tabcolsep}{8pt} 
    \begin{tabular}{p{3.2cm}p{1.5cm}p{1.5cm}} 
    \hline
    \multicolumn{1}{c}{Method} & \multicolumn{1}{c}{RMSE} & \multicolumn{1}{c}{$R^2$} \\ \hline
    \multicolumn{3}{c}{Tabular Machine Learning}                                    \\ \hline
    LightGBM Large              & 1.063                    & 0.886            \\
    KNeighborsDist             & 1.093                    & 0.879            \\
    KNeighborsUnif             & 1.096                    & 0.879            \\
    LightGBM                   & 1.108                    & 0.876            \\
    XGBoost                    & 1.124                    & 0.872            \\
    LightGBMXT                 & 1.126                    & 0.872            \\
    NeuralNetFastAI            & 1.142                    & 0.868            \\
    CatBoost                   & 1.152                    & 0.866            \\
    RandomForestMSE            & 1.182                    & 0.859            \\ \hline
    \multicolumn{3}{c}{Tabular Deep Learning}                                       \\ \hline
    TabNet                     & 0.944                    & 0.901                 \\
    ExcelFormer                & 1.001                    & 0.878                 \\
    ResNet                     & 1.014                    & 0.878                 \\
    FTTransformer              & 1.158                    & 0.850                 \\ \hline
    \multicolumn{3}{c}{Our Method}                                                 \\ \hline
    Ours                       & \textbf{0.836}            & \textbf{0.932}   \\ \hline
    \end{tabular}
\end{table}

\textbf{Experiment Setting:}  To validate the suitability of our proposed method for machine learning tasks in the Earth sciences, we conducted a comparative evaluation against a range of widely adopted machine learning baselines, including Random Forest \citep{breiman2001random}, XGBoost \citep{chen2016xgboost}, CatBoost \citep{prokhorenkova2018catboost}, the LightGBM family \citep{ke2017lightgbm}, and KNN. Additionally, we compared our method with state-of-the-art tabular deep learning approaches, including TabNet \citep{DBLP:journals/corr/abs-1908-07442}, ExcelFormer \citep{chen2024excelformerneuralnetworksurpassing}, ResNet \citep{DBLP:journals/corr/abs-2106-11959}, and FTTransformer \citep{DBLP:journals/corr/abs-2106-11959}. All models were trained using the complete dataset of 50 million samples to assess their scalability and performance on large-scale data. The prediction accuracy of each method was evaluated on the Climate2GPP test set for estimating the total gross primary productivity (GPP) for the year 2020, as detailed in the table (\ref{tab:comparison_methods}).

\textbf{Comparison:} As shown in Table \ref{tab:comparison_methods}, the best-performing machine learning and deep learning methods on this task were LightGBM Large and TabNet, achieving RMSEs of 1.063 and 0.944, respectively. However, our proposed method outperformed both, achieving a lower RMSE of 0.836 and an $R^2$ of 0.932. These results underscore the superior capability of our approach in handling large-scale Earth science data. 

\begin{figure*}[ht] 
    \centering  
    \includegraphics[width=0.96\textwidth]{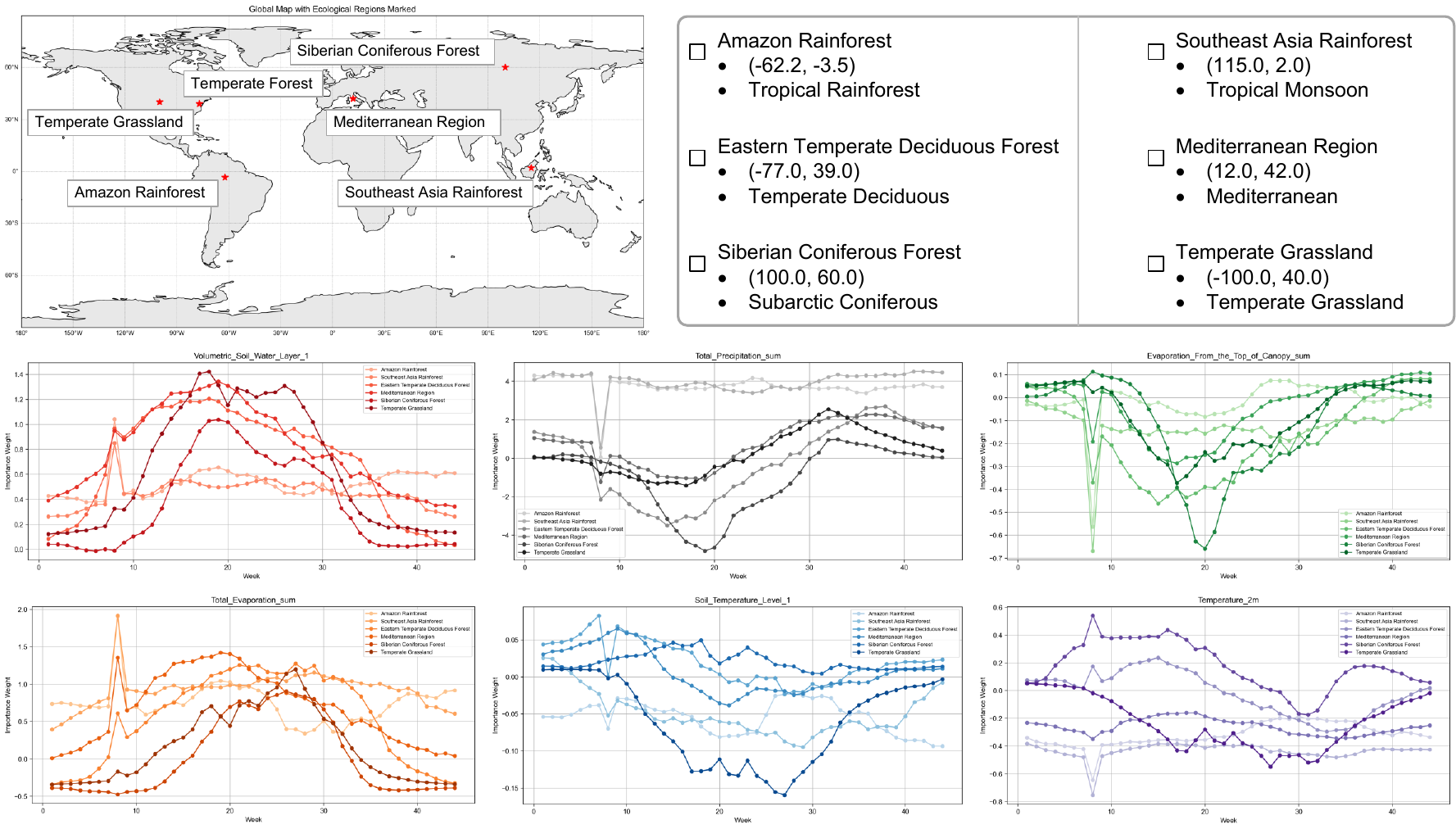}  
    \caption{Temporal Variation of Feature Importance in Six Typical Regions.}
    \label{fig:visualization} 
\end{figure*}  

\begin{figure}[ht] 
    \centering  
    \includegraphics[width=0.42\textwidth]{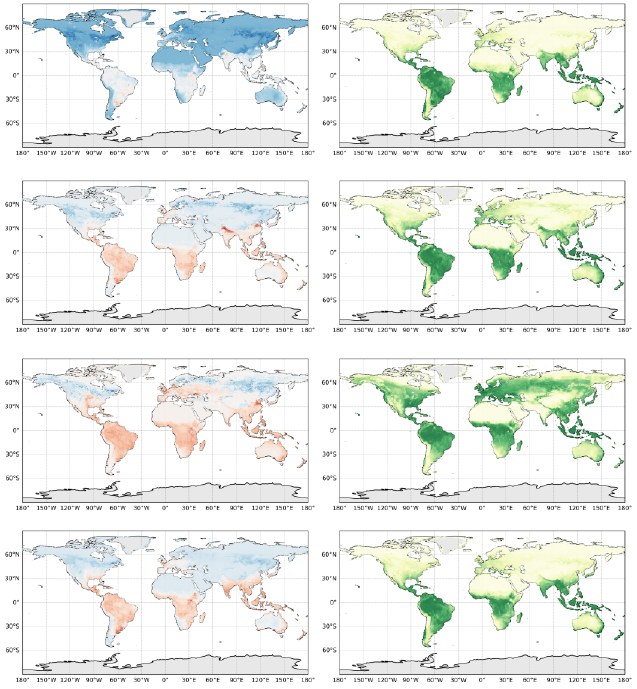}  
    \caption{Spatial Distribution of temperature\_2m Importance Weights and GPP Prediction Results at 01, 10, 20, and 30.}
    \label{fig:result} 
\end{figure}  

\subsection{Comparison of Spatial and Spatiotemporal Heterogeneity Methods}
\textbf{Experiment Setting:} Furthermore, we aim to compare our method with other approaches that are capable of modeling spatial or spatiotemporal heterogeneity. The computational complexity of the GWR series methods is proportional to the number of spatial locations in the dataset and require exactly one sample point per spatial location. Given these limitations, all experiments were conducted on a few-shot dataset. Specifically, we uniformly sampled 6,000 grids from the land grid as training data, with each grid containing data from all available time points. Since GWR series methods require a one-to-one correspondence between samples and locations, we used the average data from every 8 days over 19 years as the training samples. For GWR \citep{fotheringham2009geographically} and GNNWR \citep{du2020geographically}, which cannot model spatiotemporal heterogeneity, we fit separate weekly temporal models. For all models, we selected results from Weeks 1, 10, 20, 30, and 40, which are representative of different seasons, for comparison. The results are shown in Table \ref{tab:spatial_comparison}.

\begin{table}[h]
\centering
\caption{Comparison of Spatial and Spatiotemporal Heterogeneity Methods (RMSE / $R^{2}$)}
\label{tab:spatial_comparison}
\begin{tabular}{c@{\hspace{5pt}}c@{\hspace{5pt}}c@{\hspace{5pt}}c@{\hspace{5pt}}c}
\hline
\multicolumn{1}{c|}{} & \multicolumn{2}{c}{Spatial} & \multicolumn{2}{c}{Spatiotemporal} \\ \cline{2-5}
\multicolumn{1}{c|}{} & GWR & GNNWR & GTWR & Ours \\ \hline
\multicolumn{1}{c|}{Week-01}  & 1.990/0.43 & 0.871/0.89 & 1.761/0.56 & \textbf{0.779/0.91} \\
\multicolumn{1}{c|}{Week-10} & 2.097/0.42 & 1.066/0.85 & 1.859/0.55 & \textbf{0.813/0.91} \\
\multicolumn{1}{c|}{Week-20} & 2.184/0.59 & 1.330/0.86 & 2.475/0.48 & \textbf{1.073/0.91} \\
\multicolumn{1}{c|}{Week-30} & 2.060/0.51 & 1.178/0.84 & 2.070/0.50 & \textbf{0.931/0.89} \\
\multicolumn{1}{c|}{Week-40} & 1.958/0.43 & 0.835/0.90 & 1.689/0.58 & \textbf{0.700/0.92} \\\hline
\end{tabular}
\end{table}

\textbf{Linear Model vs non-Linear Model:} Based on the results shown in Table \ref{tab:spatial_comparison}, non-linear models such as GNNWR and our method consistently achieve better performance than the linear approaches (GWR and GTWR). For instance, in Week-01, the RMSE values for GWR and GTWR are 1.990 and 1.761, respectively, whereas GNNWR and our method obtain lower RMSE values of 0.871 and 0.779, with corresponding R² values of 0.89 and 0.91. Similar advantages are evident in Week-40, where GWR and GTWR report RMSE values of 1.958 and 1.689, compared to 0.835 and 0.700 for GNNWR and our approach, along with higher R² values (0.90 and 0.92). These numerical findings clearly illustrate that non-linear methods are more adept at capturing complex spatial and spatiotemporal variations, leading to more accurate model fitting.

\begin{table}[h]
\centering
\caption{Comparison of Spatial and Spatiotemporal Heterogeneity Methods}
\label{tab:generalization_gap}
\begin{tabular}{c|cc}
\hline
Metric & GNNWR & Ours \\ \hline
Train RMSE & 0.478 & 0.627 \\
Train $R^{2}$ & 0.931 & 0.942 \\ \hline
Test RMSE & 0.835 & 0.700 \\
Test $R^{2}$ & 0.896 & 0.922 \\ \hline
Gen. Gap & 0.357 & 0.073 \\ \hline
\end{tabular}
\end{table}

\textbf{Local learning vs Global learning:}
Our model's innovation lies in leveraging the entire spatial sample to learn the spatiotemporal conditions at each location, as opposed to the traditional local fitting approach used in GWR, GTWR, and GNNWR. This global learning strategy significantly reduces the tendency to overfit in local regions. As evidenced in Table \ref{tab:generalization_gap}, although our model exhibits a slightly higher training RMSE (0.627 versus 0.478) and a marginally improved training R² (0.942 versus 0.931) compared to GNNWR, it achieves superior performance on the test set. Specifically, our method records a lower test RMSE (0.700 compared to 0.835) and a higher test R² (0.922 versus 0.896), leading to a much smaller generalization gap (0.073 versus 0.357). This outcome confirms that employing a global learning approach effectively mitigates overfitting and enhances predictive robustness.

\subsection{Visualization}
Our visualization analysis aims to demonstrate the potential of our method in spatial analysis through three sets of results. The Figure \ref{fig:visualization} presents the temporal variation curves of explanatory variables at different spatial locations, covering six distinct climate-vegetation zones: "Amazon Rainforest", "Southeast Asia Rainforest", "Eastern Temperate Deciduous Forest", "Mediterranean Region", "Siberian Coniferous Forest", and \\"Temperate Grassland". The Figure \ref{fig:result} illustrates the spatial distribution of the explanatory weights for the temperature\_2m variable at Weeks 01, 10, 20, and 30, along with the corresponding model-predicted GPP. These visualizations are designed to highlight two main points: first, that our approach can capture the variation in dominant factors influencing GPP across different regions; and second, that the spatial patterns in the temperature\_2m weights reveal both beneficiary and affected areas under the context of global warming. It is important to note that this visualization analysis represents an initial exploration, and further work is needed to fully develop the interpretability of the model.

\section{Conclusion}
In this paper, we have identified key challenges in geographical machine learning, notably the difficulty of capturing spatiotemporal heterogeneity. While traditional statistical learning methods fail to model these variations, existing approaches such as the GWR family can capture spatiotemporal heterogeneity but rely on locally sampled data, which often leads to overfitting in spatial regions.

To address these issues, we have proposed a novel deep learning–based framework that decomposes a geographical regression task into learning location-independent data features and capturing spatiotemporal variations. Unlike conventional methods, our approach optimizes over the entire data space, significantly alleviating overfitting in local areas.

We validated our method on a large-scale dataset built from ERA5 and PML\_V2, comprising 50 million training samples and 2.8 million test samples. The experimental results demonstrate the effectiveness of our design: our method achieved a test RMSE of 0.836, outperforming GWR (RMSE 1.937), classical tabular machine learning methods like LightGBM Large (RMSE 1.063), and other deep learning approaches such as TabNet (RMSE 0.944). These findings confirm that our proposed framework not only captures spatiotemporal heterogeneity effectively, but also delivers superior predictive performance in geographical machine learning tasks.

\section{Discussion}
While our proposed method shows promising performance, several aspects remain open for future improvement. First, regarding interpretability, our visualizations indicate that our approach holds potential for explaining complex spatiotemporal relationships. However, to accurately characterize causal effects, we plan to integrate additional causal constraints and develop methods that infer dominant factors directly from the explanatory weights. Second, the current modeling of spatial interactions relies on a simple linear mechanism using graph convolutions, where interaction patterns are predetermined and fixed. In future work, we intend to explore more sophisticated and dynamic spatial interaction models, including refined embedding techniques for spatiotemporal condition graphs and improved strategies for handling multimodal information during prediction. Lastly, our model has so far been validated only on the Climate2GPP dataset; additionally, we plan to test our method on simulated datasets and other real-world tasks to further verify its generalizability and robustness.

{
	\begin{spacing}{1.17}
		\normalsize
		\bibliography{ref} 
	\end{spacing}
}

\end{document}